\documentclass{article}

\PassOptionsToPackage{numbers, compress}{natbib}

\usepackage{textcomp}

  \usepackage[neuripsworkshop]{neurips_2019}

\usepackage{xr-hyper}
\usepackage[utf8]{inputenc} 
\usepackage[T1]{fontenc}    
\usepackage{hyperref}       
\usepackage{url}            
\usepackage{booktabs}       
\usepackage{amsfonts}       
\usepackage{nicefrac}       
\usepackage{microtype}      
\usepackage{soul,color}
\usepackage{bm}
\usepackage{amsmath}
\usepackage{graphicx}
\usepackage{subcaption}
\usepackage{placeins}
\usepackage{comment}
\newsavebox{\imagebox}

\DeclareMathOperator*{\argmax}{arg\,max}

\RequirePackage{algorithm}
\RequirePackage{algorithmic}

\title{Thompson Sampling for Contextual Bandit Problems with Auxiliary Safety Constraints}

\author{%
  Samuel Daulton, Shaun Singh, Vashist Avadhanula, Drew Dimmery, Eytan Bakshy\\
  Facebook\\
  Menlo Park, CA 94025 \\
  \texttt{\{sdaulton, shaundsingh, vas1089, drewd, ebakshy\}@fb.com} \\
}

\begin{document}
\maketitle
\begin{abstract}
Recent advances in contextual bandit optimization and reinforcement learning have garnered interest in applying these methods to real-world sequential decision making problems \citep{yauney18, lazic18, tang13, bottou13}. Real-world applications frequently have constraints with respect to a currently deployed policy. Many of the existing constraint-aware algorithms consider problems with a single objective (the reward) and a constraint on the reward with respect to a baseline policy. However, many important applications involve multiple competing objectives and auxiliary constraints. In this paper, we propose a novel Thompson sampling algorithm for multi-outcome contextual bandit problems with auxiliary constraints. We empirically evaluate our algorithm on a synthetic problem. Lastly, we apply our method to a real world video transcoding problem and provide a practical way for navigating the trade-off between safety and performance using Bayesian optimization.
\end{abstract}

\section{Introduction}
 In a typical contextual bandit problem, the decision maker observes a context-feature vector $\bm x \in \mathbb R^d$ and picks an action $a$ from a set of actions $\mathcal A$. The decision maker then observes a parametric reward $f_{\bm \theta}(a)$ corresponding to the selected action, where $\bm \theta \in \mathbb R^{d} $ is unknown apriori. The objective is to maximize the cumulative reward from the selected actions over a time-horizon,  while simultaneously learning $\bm \theta$. The contextual bandit framework is commonly used to balance the exploration-exploitation trade-offs involved in high-dimensional dynamic decision making, including personalized recommendation \citep{li2010} and clinical trials \citep{bastani2015online, villar2018covariate}.  While contextual bandits are well studied in the context of optimizing a single reward metric, there is little understanding of deploying contextual bandits in settings involving multiple objectives (metrics), which often compete with each other. 
 
 In this work, we focus on developing practical bandit approaches that can manage both the exploration-exploitation trade-off, as well as the trade-offs associated with optimizing multiple metrics. In many applications, the decision maker is concerned with improving multiple metrics under uncertain information. Furthermore, it is common that some of the metrics compete with each other: frequently, improvement in one metric comes at a cost in another metric. Therefore, using contextual bandit algorithms that optimize for a single objective function is not desirable. For example, consider an example of uploading videos, an application of particular interest to internet platforms and the focus of this paper. When a user decides to share a video via an online platform, the platform determines what quality (up to the quality at which the video was recorded) at which to upload the video. With the explosion of demand for video content, platforms seek to enable users to share videos of increasingly high quality without reducing upload reliability. These goals, however, are in conflict: increasing quality typically makes sharing \emph{less} reliable. As the video's file size increases, upload time increases, making it more likely that transient connectivity will lead to failure or that an impatient user will cancel the upload. As such, the demands of improving a video upload service dictate that increases in quality must occur \emph{without} hurting the reliability of the service. In other words, the goal of optimization is, for each state, to find the action leading to the highest quality among actions which will not reduce upload reliability. This set of actions may be different for each user.

We consider a stylized contextual bandit problem with auxiliary safety constraints that shares salient features with the multi-objective bandit problem, where the goal is to develop an efficient exploration-exploitation policy that can simultaneously manage trade-offs associated with optimizing multiple metrics. In particular, we focus on the two-metric setting and formulate a constrained contextual bandits problem, where we consider one of the metrics as the objective (reward) to optimize while enforcing a safety constraint on the other (auxiliary) metric with respect to a baseline ``status quo'' policy.\footnote{Our approach generalizes to multiple auxiliary safety constraints on different metrics, but we focus to the case of a single auxiliary safety constraint and metric for brevity.} We refer to this constraint as the \textit{auxiliary safety constraint}. The underlying assumption is that the decision maker is interested in optimizing the first metric only when the second metric is not going to be degraded ``too much'' relative to the current status quo. This is a reasonable assumption in many settings and in particular the video uploads application described earlier, where the platform is focused on improving the quality without hurting the reliability. In this paper, we present a simple and robust optimization procedure based on Thompson sampling (TS) which meets this criteria.

\section{Related Work}
Recently, contextual bandits have attracted increased attention. The works of \citep{auer02, dani2008, li2010, chu11, agrawal13,  abbasi13} have considered the linear contextual bandits, a popular variant of the contextual bandits problem and established strong theoretical guarantees while demonstrating empirical efficacy. More recently, \citet{filippi13} has considered the generalized contextual bandit problem and developed an upper-confidence bound (UCB) based algorithm. However, the focus of the stream of work is on optimizing a single objective and cannot be easily adapted to balance trade-offs associated with optimizing multiple objectives. 

Our work is more closely related the constrained contextual bandit framework. \citet{badanidiyuru14} consider the linear bandit problem with linear budget constraints and present a primal-dual algorithm that uses multiplicative updates to achieve a sub-linear regret. \citet{agrawal2014bandits} consider a generalized version of this problem, where the objective is concave and constraints are convex and presents a UCB-based algorithm with regret guarantees. The constraints in both these papers are modeled on the lines of resource (or budget) limitations and both the works assume that the constraint is deterministic and known apriori. In contrast, the constraint in our problem is stochastic and not known apriori. In particular, the rewards under the baseline policy depend on the time-varying contexts. Therefore, these two works do not trivially generalize to the ``bandits with safety constraints.'' The works of \citet{amani2019linear} and \citet{kazerouni17} consider the linear contextual bandit problem with ``safety constraints'' and present UCB-based approaches that ensure that the new policy does not violate the ``safety constraint.'' While the objectives of both these works are similar, neither directly translates to the contextual bandit problem with auxiliary safety constraints; \citet{amani2019linear} assume the bound in the safety constraint is deterministic and known apriori for every round, and the algorithms by \citet{kazerouni17} address the single objective setting, where the focus is on ensuring that the aggressive exploration tendencies of UCB approach do not hurt the objective in the early rounds in comparison to the baseline policy. 

In this paper, we present a simple and robust optimization procedure based on Thompson sampling (TS) which meets this criteria. This is primarily motivated by the attractive empirical properties of TS-based approaches for bandit problems that have been observed in a stream of recent papers \citep{chapelle11, agrawal2017thompson, riquelme18}. One of the main contributions of this work is in highlighting the salient features of the constrained contextual bandits problem that need to be adapted or customized to facilitate the design of a TS algorithm for the multi-objective bandit problem. To the best of our knowledge, this is the first attempt in designing policies for contextual bandit problems with auxiliary safety constraints and can hopefully serve as a stepping stone in designing robust multi-objective bandit algorithms with theoretical guarantees.

\section{Contextual Bandits with Auxiliary Safety Constraints}

\subsection{Problem Formulation}
In the contextual bandit setting with an auxiliary constraint, at any time $t$, an agent is presented with a context vector, $\mathbf x_t \in \mathbb{R}^d$. The agent selects an action $a_t$, from a possibly infinite and time-dependent set, $\mathcal A_t$, and observes two potentially correlated metrics: a reward metric $r_t \in \mathbb R$ and constraint metric $c_t \in \mathbb R^+$ sampled from distributions (models) $f^{(r)}(\bm x, a)$ and $f^{(c)}(\bm x, a)$ respectively. In addition, at each time step $t$ the agent queries a baseline policy $\pi_b$ and receives the action selected by the baseline policy $b_t = \pi_b(\bm x_t)$. For a given value of $\alpha$, the goal is to maximize cumulative reward $$R_t = \sum_{t=1}^T r_t,$$ while simultaneously learning a policy that satisfies the constraint $$\mathbb E[c_t|\bm x_t, a_t] \geq (1-\alpha)\mathbb E[c_t|\bm x_t, b_t]$$ at each time step $t$. In this setting, $\alpha$ can be interpreted as the maximum decrease the decision maker is willing to accept in the constraint metric, while optimizing the reward metric. Though it might be intuitive to apply the constraint cumulatively as in \citet{kazerouni17}, we focus on enforcing the constraint for each instance to ensure fairness; using an aggregate constraint could lead to a policy that degrades the constraint metric for certain sub-populations. For example, in the case of video uploads, maximizing quality while enforcing an aggregate constraint on reliability could lead to a policy that has no change in reliability for uploading low quality videos, but reduces reliability for uploading high quality videos. 

\subsection{Thompson Sampling with Auxiliary Safety Constraints (TS-ASC)}
We propose a novel algorithm called Thompson Sampling with Auxiliary Safety Constraints (TS-ASC) for policy learning in contextual bandit problems with auxiliary safety constraints. The algorithm, like any TS-based approach, begins with an initial distribution over models $\mathbb P(\bm f)$, where $\bm f$ is a model that outputs an estimated reward metric $f^{(r)}(\bm x, a)$ and an estimated constraint metric $f^{(c)}(\bm x, a)$.\footnote{The reward and the constraint could be modeled independently or jointly, but for brevity we denote the joint or independent models together by $\bm f$.}  At each time step $t$, a model is sampled from the posterior distribution over models $\bm \tilde{f}_t \sim \mathbb P(\bm f|\mathcal D_t)$ given the observed data $\mathcal D_t = \{(\bm x_i, a_i, r_i, c_i)\}_{i=1}^{t-1}$. Next, to ensure that the action picked by the TS-ASC satisfies the auxiliary safety constraint, we need to compare the constraint metric value obtained from the TS-ASC policy to the constraint metric value obtained from the baseline policy. To obtain the constraint value under the baseline policy, we approximate the value of the constraint using the model sampled from the posterior. Concretely, in each round a set of feasible actions is defined to contain all actions $a \in \mathcal A$ that satisfy the constraint under the sampled model $\bm \tilde{f}_t$; that is, $$\mathcal A_{\text{feas}} \leftarrow \{a \in \mathcal A | \tilde{f}^{(c)}_t(\bm x_t, a) \geq (1-\alpha) \tilde{f}^{(c)}_t(\bm x_t, b_t)\}.$$ The policy TS-ASC selects the action $a$ with maximum estimated reward $\tilde{f}^{(r)}_t(\bm x_t, a)$ under the sampled model. We provide the details of TS-ASC in Algorithm \ref{alg:constrained_thompson_sampling}.

\medskip \noindent {\bf Remark:} While, the accuracy of the constraint typically depends on the accuracy of the sampled model, we observe in our numerical simulations that this relaxation does not degrade the empirical performance with regards to constraint violation. We believe that as long as there is correlation between the reward and constraint metrics, the posterior updates of TS will eventually lead to estimating the constraint value accurately. Proving such a statement theoretically will most likely involve a notion of correlation between metrics and should be an interesting future direction. 

\begin{algorithm}[ht]
   \caption{TS-ASC}
   \label{alg:constrained_thompson_sampling}
\begin{algorithmic}[1]
    \STATE {\bfseries Input:} set of actions $\mathcal A$, initial distribution over models $P(f)$, $\mathcal D_1 = \emptyset$
    \FOR{$t=1$ {\bfseries to} $T$}
        \STATE Receive context $\bm x_t$
        \STATE Sample model $\bm \tilde{f}_t \sim \mathbb P(\bm f|\mathcal D_t)$
        \STATE Query action from baseline policy $b_t \leftarrow \pi_b(\bm x_t)$
       \STATE Determine feasible actions: $\mathcal A_{\text{feas}} \leftarrow \{a \in \mathcal A | \tilde{f}^{(c)}_t(\bm x_t, a) \geq (1-\alpha) \tilde{f}^{(c)}_t(\bm x_t, b_t)\}$
       \STATE Select an action $a_t \leftarrow \argmax_{a \in \mathcal A_{\text{feas}}} \tilde{f}^{(r)}_t(\bm x_t, a)$
       \STATE Observe outcomes $r_t, c_t$
       \STATE $\mathcal D_{t+1} \leftarrow \mathcal D_t \cup \{\bm (x_t, a_t, r_t, c_t)\}$
   \ENDFOR
\end{algorithmic}
\end{algorithm}

\section{Synthetic Experiment}
\subsection{Problem Setup}
To demonstrate the efficacy of the TS-ASC algorithm, we adapt the synthetic example proposed by \citet{kazerouni17}. The problem consists of 100 arms, each with a corresponding feature vector $\mathbf x_a \in \mathbb{R}^4$. The reward metric $r_t \sim \mathcal N(\mathbf x_a^T\bm \theta^{(r)},\sigma^2)$ and the constraint metric $c_t \sim \mathcal N(\mathbf x_a^T\bm \theta^{(c)},\sigma^2)$ are sampled independently from two different normal distributions with the respective means being a linear functions of the feature vectors and a known noise standard deviation $\sigma=0.1$. We define the baseline policy $\pi_b$ to be a fixed policy that takes the same action $b$ at each time step $t$. To set the baseline action $b$, we select the top 30 actions with respect their expected reward, sort those 30 actions with respect to their expected constraint metric, and set $b$ to be the $20^{\text{th}}$ best action with respect to the expected constraint metric. The goal is to maximize cumulative reward,  $\sum_{t=1}^{T} r_t$, while learning a policy that satisfies the constraint $\mathbb E[c_t|a_t] \geq (1-\alpha)\mathbb E[c_t|b_t]$ with high probability. Following \citet{kazerouni17}, the true parameters $\bm \theta^{(r)}, \bm \theta^{(c)}$ and the feature vectors are all $iid$ samples from $\mathcal N(0,I_4)$ such that the expected values of each metric $x_a^T\bm \theta_1, x_a^T\bm \theta_2$ are positive for all actions. Additionally, we construct the problem that the maximum expected reward among feasible actions is less than the maximum reward among infeasible actions in order to create a trade-off between constraint satisfaction and cumulative reward. We model each outcome using an independent ridge regression model (using the Bayesian interpretation for TS) with $\lambda=1$. We follow \citet{kazerouni17} and set $\delta= 0.001$ as is done in the original CLUCB synthetic experiment. 
\subsection{Baseline Algorithm}
To the best of our knowledge, there are no existing algorithms for addressing contextual bandit problems with auxiliary safety constraints. Since the Conservative Linear Upper Confidence Bound (CLUCB2) algorithm \citep{kazerouni17} is the closest related work to this paper, we extend CLUCB2 to our problem for comparison. The original CLUCB2 algorithm selects the optimistic action $a$ only if a performance constraint with respect to a baseline policy is satisfied with high probability and otherwise takes the action selected by the baseline policy. The performance constraint used by CLUCB2 requires that the cumulative reward of the proposed policy be greater than some fraction $\alpha$ of the cumulative reward for the baseline policy with high probability at all time steps $t$. More specifically,
$$\sum_{i=1}^t \hat{r}_{a_i}^i \leq (1-\alpha) \sum_{i=1}^t \hat{r}_{b_i}^i.$$
We modify this algorithm by moving the constraint from the reward metric to constraint metric. Furthermore, we consider two variations of the the adapted CLUCB2 algorithm: one variant, referred to as  CLUCB2-ASC-C, follows the original CLUCB2 and evaluates the constraint cumulatively and the other variant, referred to as  CLUCB2-ASC-I, evaluates the constraint at the instance-level rather than enforcing it cumulatively. Additionally, at each time step $t$, our both CLUCB2-ASC-C and CLUCB2-ASC-I create a set of feasible actions $\mathcal A_{\text{feas}}$ that satisfy the auxiliary safety constraint and select the actions from the feasible set with the maximum upper confidence bound on the reward metric. We provide the exact details for CLUCB2-ASC-C in Algorithm \ref{alg:clucb2_asc}. 

\begin{algorithm}[ht]
   \caption{CLUCB2-ASC-C}
   \label{alg:clucb2_asc}
\begin{algorithmic}[1]
    \STATE {\bfseries Input:} discrete set of actions $\mathcal A, \alpha, r_l, \mathcal B, \mathcal F$
    \STATE {\bfseries Initialize:} $n \leftarrow 0$, $\bm z \leftarrow \bm 0$, $\bm w \leftarrow \bm 0$, $\bm v \leftarrow \bm 0$, $\mathcal C_1^{(r)} \leftarrow \mathcal B, \mathcal C_1^{(c)} \leftarrow \mathcal B$
    \FOR{$t=1$ {\bfseries to} $T$}
        \STATE Receive context $\bm x_t$
        \STATE Query action from baseline policy $b_t \leftarrow \pi_b(\bm x_t)$
        \STATE $\mathcal A_{\text{feas}} \leftarrow \{b_t\}$
        \FOR{$a'_t \in \mathcal A$}
           \STATE Compute: $R_t \leftarrow \max_{\bm \theta^{(c)}\in \mathcal C_t^{(c)}} (\bm v + \bm x_{b_t})^T\bm \theta^{(c)}$
           \STATE Compute: $L_t \leftarrow \min_{\bm \theta^{(c)}\in \mathcal C_t^{(c)}} (\bm z + \bm x_{a'_t})^T\bm \theta^{(c)} + \alpha \max \big(\min_{\bm \theta^{(g)}\in \mathcal C_t^{(c)}} \bm w^T\bm \theta^{(c)}, nr_l\big)$
           \IF{$L_t \geq (1-\alpha)R_t$}
                \STATE $\mathcal A_{\text{feas}} \leftarrow \mathcal A_{\text{feas}} \cup \{a'_t\}$
           \ENDIF
       \ENDFOR
       \STATE Select optimistic action: $(a_t, \tilde{\bm \theta}^{(r)}) \leftarrow \argmax_{(a_t, \bm \theta^{(r)}) \in \mathcal A_{\text{feas}} \times \mathcal C_t^{(r)}} \bm x_{at}^T \bm \theta^{(r)}$
       \IF{$a_t \ne b_t$}
                
            \STATE $\bm z \leftarrow \bm z + \bm x_{a'_t}$
            \STATE $\bm v \leftarrow \bm v + \bm x_{b_t}$
        \ELSE
            \STATE $\bm w \leftarrow \bm w + \bm x_{b_t}$
            \STATE $n \leftarrow n + 1$
       \ENDIF
       
       \STATE Take action $a_t$ and observe outcomes $r_t, c_t$
       \STATE Update confidence sets $\mathcal C_{t+1}^{(r)}, \mathcal C_{t+1}^{(c)}$ using $(\bm x_t, a_t, c_t, y_t)$
   \ENDFOR
\end{algorithmic}
\end{algorithm}
\begin{figure}[ht]
    \centering
    \begin{subfigure}{\textwidth}
        \includegraphics[width=\textwidth]{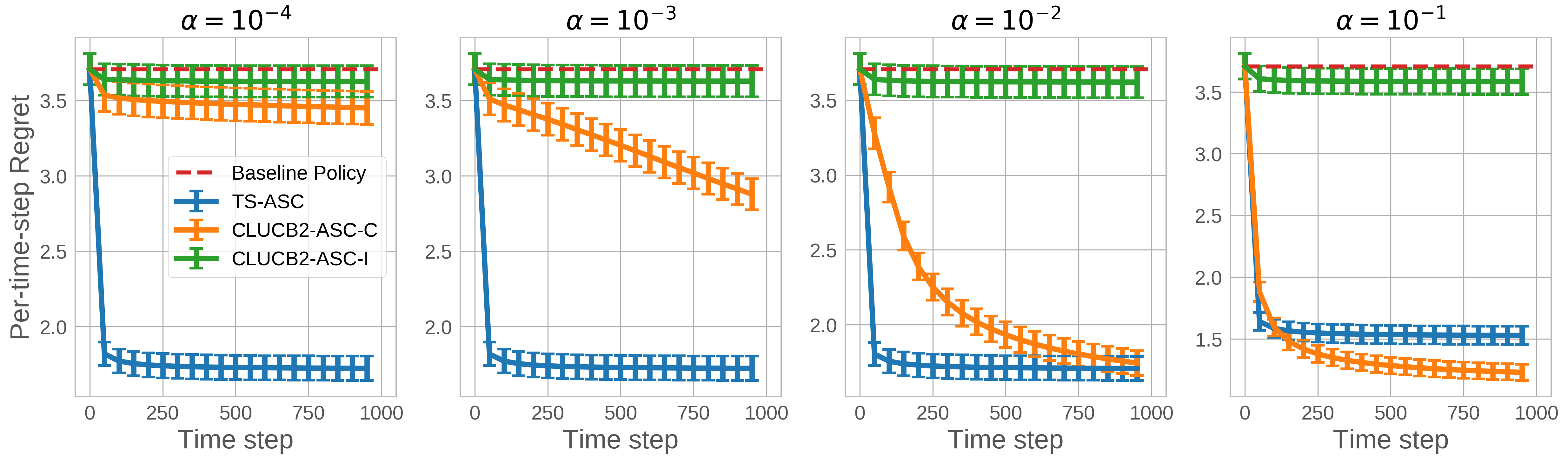}
        \subcaption{\label{fig:toy_results:regret} Per-time-step Regret}
    \end{subfigure} %
    \vskip\baselineskip
    \begin{subfigure}{\textwidth}
        \includegraphics[width=\textwidth]{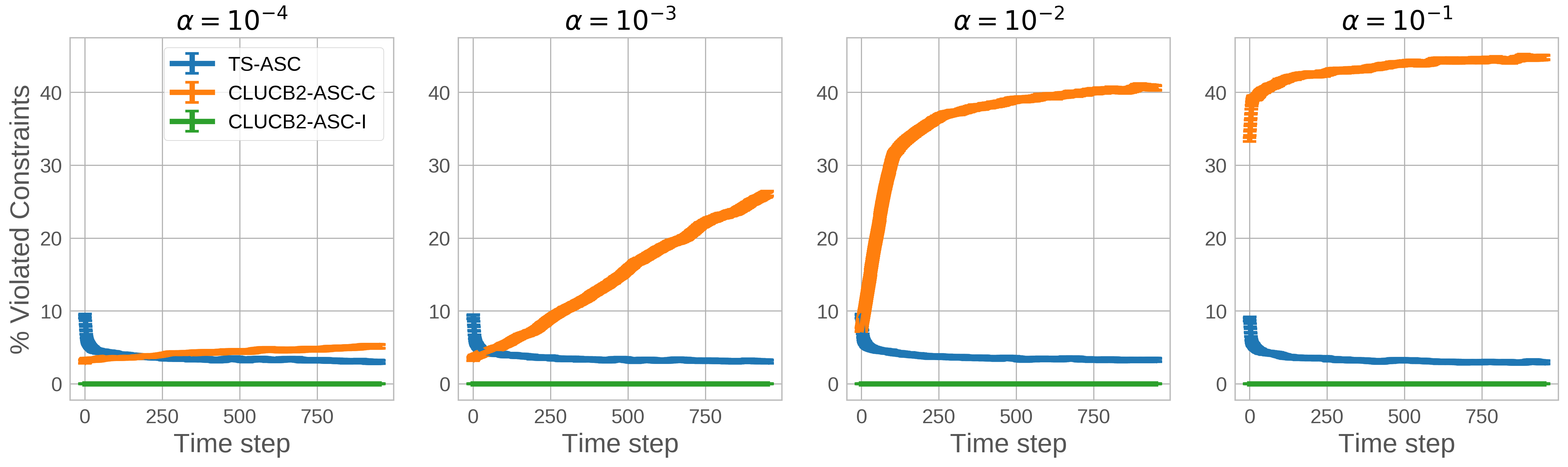}
        \subcaption{\label{fig:toy_results:pct_violated} Moving average of the percentage of instance-level constraints that are violated.}
    \end{subfigure}
    \caption{\label{fig:toy_results} Evaluation of constrained contextual bandit algorithms in a synthetic problem.}
\end{figure}

\subsection{Results}
We evaluate TS-ASC across 1000 realizations of the synthetic problem and compare TS-ASC, CLUCB2-ASC-C, and CLUCB2-ASC-I. Figure \ref{fig:toy_results} plots the mean and 2 standard errors of the mean for the evaluation metrics of interest. Figure \ref{fig:toy_results:regret} shows the per-time-step regret for all of the algorithms for different values of $\alpha$. For all values of $\alpha$, TS-ASC and CLUCB2-ASC-C achieve far lower regret compared to the baseline policy, where as CLUCB2-ASC-I is extremely conservative and does deviate far from the baseline policy. However, there are significant differences in the policies learned by CLUCB2-ASC-C and TS-ASC. Figure \ref{fig:toy_results:pct_violated} shows that TS-ASC ultimately learns a policy that satisfies the instance-level constraint with high probability, where as CLUCB2-ASC-C learns a policy that violates the instance-level constraint more than 40\% of the time for larger values of $\alpha$. This is because the cumulative performance constraint used in CLUCB2-ASC-C does not require constraint satisfaction at the instance-level with high probability. While CLUCB2-ASC-I, does always satisfies the instance-level constraint, it is very conservative and has significantly higher regret than TS-ASC. That is, TS-ASC navigates an effective trade-off between regret and constraint satisfaction relative to the CLUCB2-ASC methods. It does this with only slightly more common constraint violation early in the learning process (as it explores slightly more). In doing so, it is able to effectively avoid constraint violation in the final learned policy, unlike CLUCB2-ASC-I.
\subsection{Fairness}
\begin{figure}
\centering
\savebox{\imagebox}{\includegraphics[width=.3\textwidth]{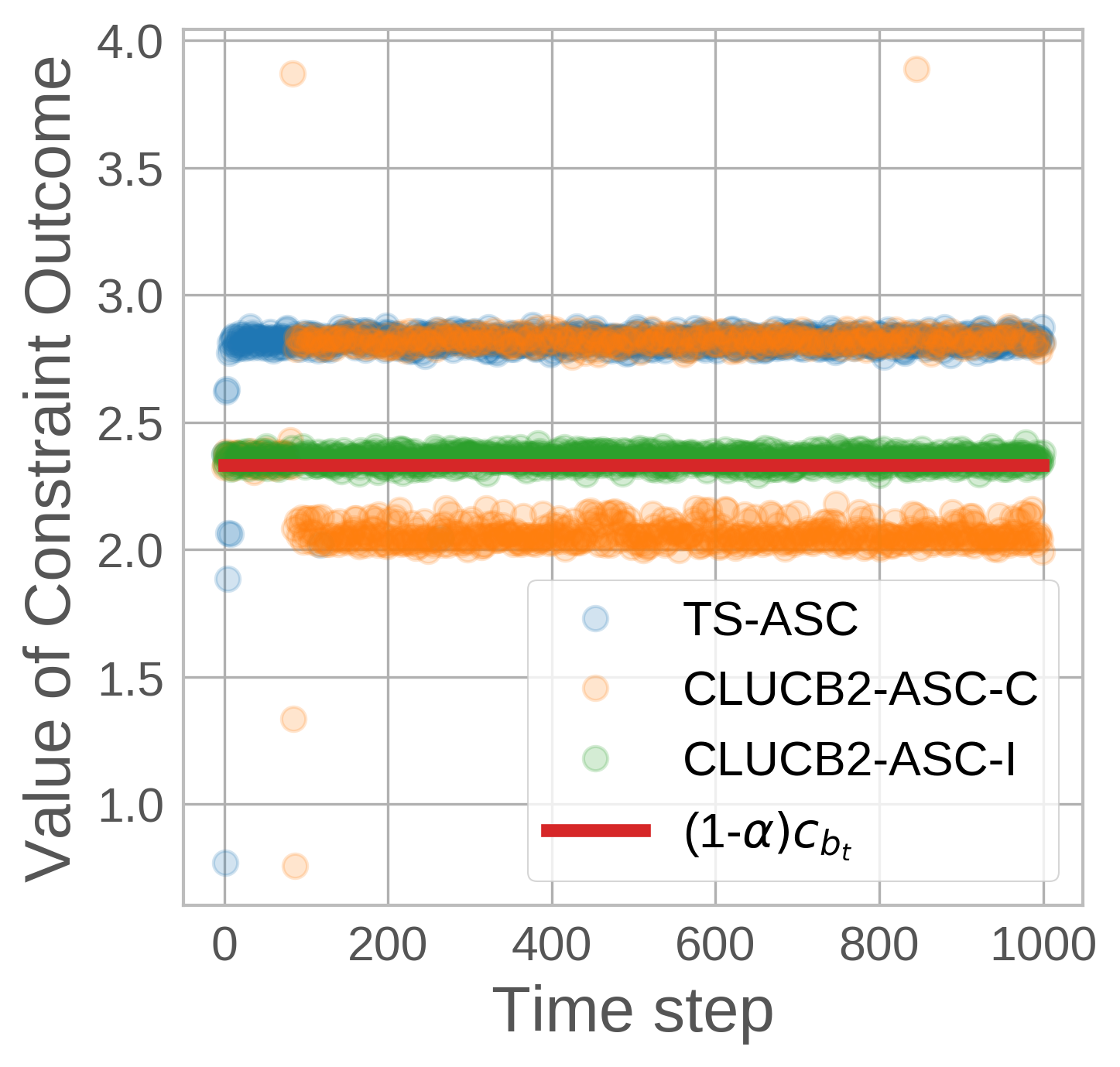}}%
  \begin{subfigure}[t]{.45\linewidth}
    \centering\usebox{\imagebox}
    \caption{\label{fig:constraint_outcomes} The expected constraint outcome $\mathbb E[c_t|\bm x_t, a_t]$ under each policy along with the constraint threshold: $(1-\alpha)\mathbb E[c_t|\bm x_t, b_t]$. A small amount of zero-mean, iid Gaussian noise with standard deviation of 0.02 has been added to improve visibility. This figure shows data from a single realization of the problem using $\alpha = 10^{-2}$.}
  \end{subfigure}\qquad
  \begin{subtable}[t]{.45\linewidth}
    \centering\raisebox{\dimexpr.5\ht\imagebox-.5\height}{
      \begin{tabular}{||c|c|c||}
        \hline
        $\alpha$     & Mean & SEM\\
        \hline\hline
        $10^{-1}$ & 1.2181 & 0.0097 \\
        $10^{-2}$     & 1.2980 &0.0097 \\
        $10^{-3}$     & 1.3065       &0.0096\\
        $10^{-4}$     & 1.3077 & 0.0096       \\
        \hline
      \end{tabular}
    }%
    \caption{\label{fig:table}The expected constraint value (normalized with respect to the expected baseline constraint value) under the TS-ASC policy over the last 100 time steps for various values of $\alpha$. The means and standard errors over 1000 realizations are reported.}
  \end{subtable}
  \caption{Evaluation of Constraint values}
\end{figure}

The cumulative performance constraint used by CLUCB2-ASC-C leaves room for the policy to select a mixture of actions with low values of the constraint metric $c_t$ (which would violate an instance level constraint) and high values of the constraint metric $c_t$ such that the cumulative constraint is still satisfied with high probability. Figure \ref{fig:constraint_outcomes} highlights the difference in learned policy. The red line indicates the instance-level constraint threshold $(1-\alpha)\mathbb E[c_t(\bm x_t, b_t)]$. For each policy, the point at time step $t$ is the expected constraint value $\mathbb E[c_t| \bm x_t, a_t$ under the policy. Hence, points above the red line satisfy the instance-level constraint and points below the red line violate the instance-level constraint. TS-ASC quickly converges to a policy that takes a single action that satisifies the instance-level constraint, where CLUCB2-ASC-C ultimately converges to a policy selects a mixture of infeasible and feasible actions with respect to the instance-level constraint. While the CLUCB2-ASC-C achieves lower regret than the TS-ASC policy, it comes at the cost of a high rate of instance-level constraint violations. Importantly, we choose to use an instance-level constraint in the problem specification and in our evaluation because it enforces fairness: that is, we want policies that satisfy the constraint at the instance-level, not just the global level. Of course, if the baseline policy is not fair, then this constraint will merely perpetuate those inequities, although it will, at least, not exacerbate them by more than $\alpha$.
\subsection{Choosing $\alpha$}
Figure \ref{fig:toy_results} highlights a trade-off in regret and constraint satisfaction. While TS-ASC achieves significantly lower regret than the baseline policy and has a lower rate of instance-level constraint violations than CLUCB2-ASC-C, TS-ASC occasionally violates the instance-level constraint for all evaluated values of $\alpha$ due to randomness associated with posterior sampling. Importantly, the choice of $\alpha$ controls not only the algorithm's constraint check within TS-ASC, but also the true instance-level constraint evaluation. Therefore, decreasing $\alpha$ means that there is a higher instance-level constraint value threshold that TS-ASC must meet. As shown in Table \ref{fig:table}, smaller values of $\alpha$ will lead to policies with larger instance-level constraint values. Depending on the tolerance for violated constraints, the value of $\alpha$ used in the problem specification may not be same value of $\alpha$ that should be used by TS-ASC achieve the desired constraint outcome. In some cases, it may be desired to make the constraint criterion in the TS-ASC algorithm stricter than the constraint of the problem specification in order reduce the likelihood of taking infeasible actions. In addition, the trade-off between the reward and constraint metrics may not be obvious in many cases. In Section \ref{sec:reward_shaping}, we demonstrate how the $\alpha$ can be tuned using Bayesian optimization in order to achieve a satisfactory trade-off between the reward and constraint metrics Section \ref{sec:video_uploads}.

\section{Video Upload Transcoding}
\label{sec:video_uploads}
We demonstrate the utility of TS-ASC on a real world video upload transcoding application serving millions of uploads per day. The task is sequential decision making problem, where the policy receives a request that a video be uploaded and the policy must choose the upload quality (or equivalently how to transcode the video) given contextual features about the video. The context $\bm x$ includes information such as device model and operating system, connection class (e.g. 2G, 3G, etc.), network download speed, and the country, as well as features about the video such as the file size, the source bitrate, and the resolution. Uploading at a higher quality is desired, but is less likely to succeed than uploading at a lower quality because the upload takes longer to complete, which increases the likelihood that a frustrated user cancels the upload or experiences a connectivity failure. The high level goal is to maximize quality among successfully uploaded videos without reducing upload reliability relative to a baseline policy. It is important to note that the video quality is deterministic and known apriori given the upload quality---the only stochasticity is around whether the upload is successful or not. For example, if a video with a source quality of 720p is trancoded to 480p, it is known in advance that if the upload succeeds, the video quality will be 480p.

\subsection{Problem Formulation}
We define the task as a contextual bandit problem with an auxiliary safety constraint, where the agent receives a context $\bm x \in \mathbb R^{39}$ and selects one of 4 actions in $\mathcal A = \{\text{360p}, \text{480p}, \text{720p}, \text{1080p}\}$. As in many applications, the true reward function is unknown, and it is not obvious how to specify the reward function to achieve the goal of increasing video upload quality without decreasing reliability. We choose a tabular reward function where the reward is 0 for a failed upload, and the reward strictly, monotonically increases for successful uploads with higher qualities. We parameterize the reward function in terms of a set of positive, additive offsets $\Omega = \{\omega_a~\forall ~ a\in\mathcal A\}$ such that $r_t = \mathbb I(y_t = 1)\sum_{a' \leq a_t} \omega_{a'}$. To ensure high reliability relative to the baseline production policy $\pi_b$, we use the success outcome $y_t|\bm x_t, a_t$ as the constraint metric and introduce an auxiliary safety constraint on reliability: $\mathbb E[c_t|\bm x_t, a_t] \geq (1-\alpha)\mathbb E[c_t|\bm x_t, b_t]$, where $b_t = \pi_b(\bm x_t)$. Importantly, this choice of instance-level auxiliary safety constraint encodes a fairness requirement: namely, it requires that the probability of a successful upload (the expected value of the Bernoulli success outcome) under the new policy must be at least $1-\alpha$ times the probability of a successful upload under the baseline policy$\pi_b$. Hence, the auxiliary safety constraint protects against policies that achieve satisfactory reliability at the global-level through a mixture of high and low reliability actions, which could lead to a disproportionately poorer upload experience for sub-populations of videos, networks, or devices relative to the status quo.

\subsection{Modeling Approach}
Since the reward is deterministic given the success outcome $y_t$ and the action $a$, we only need to model success outcome $y_t$ as a function of the context $\bm x_t$ and action $a_t$. Since the task includes several sparse features, we follow recent work on using Bayesian deep learning for Thompson sampling \citep{riquelme18}. For this task, we leverage ideas from deep kernel learning \citep{wilson16} and NeuralLinear models \citep{riquelme18} in using a shared two-layer fully-connected neural network (with 50 and 4 hidden units respectively) as feature extractor to learn a feature representation $\bm \phi_a$ for each context-action feature vector $\mathbf x_a$. We use the learned feature representation as the input to a Gaussian Process (GP) with a linear kernel and a Bernoulli likelihood. For details regarding the modeling and training procedure, see Appendix \ref{app:modeling}.

\subsection{Reward Shaping and Tuning $\alpha$}
\label{sec:reward_shaping}
Since the optimal parameters $\Omega$ of the reward function and value of $\alpha$ are unknown, we use Bayesian optimization to jointly optimize the reward function \citep{reward-shaping} and $\alpha$ to find parameters that achieve a satisfactory trade-off between metrics. While there many ways to measure the quality of video uploads, our strictly increasing reward function with respect quality and reliability constraint create correlations between many of these quality metrics. We select one quality metric for our objective---the fraction of 1080p videos that are uploaded at 1080p, which we refer to as the quality preserved at 1080p---and report all quality metrics to the system designers for evaluation. Formally, the objective is $$\max_{\alpha, \Omega} \mathbb E\big[q_{\text{1080p}}|\pi(\alpha, \Omega)\big] ~~\text{s.t.}~~\mathbb E\big[c|\pi(\alpha, \Omega)\big] \geq \mathbb E\big[c|\pi_b\big],$$ where $q_{\text{1080p}}$ is a binary variable indicating if the 1080p upload quality is preserved and $\pi(\alpha, \Omega)$ is the TS-ASC policy learned given $\alpha$ and  $\Omega$. We fix $\omega_{\text{360p}}=1$ and $\omega_{\text{480p}} \in (0, 0.05],~ \omega_{\text{720p}} \in (0, 0.04],~ \omega_{\text{1080p}} \in (0, 0.03]$ and $\alpha \in [0, 0.06]$. Each parameterization is evaluated using an A/B test which allows us to measure the mean and standard error of each metric. Since the optimization problem involves noisy outcomes and outcome constraints, we use Noisy Expected Improvement \citep{letham2019} as our acquisition function.

\subsection{Results}
\begin{figure}[ht]
    \centering
    \begin{subfigure}{.33\textwidth}
        \includegraphics[width=\columnwidth]{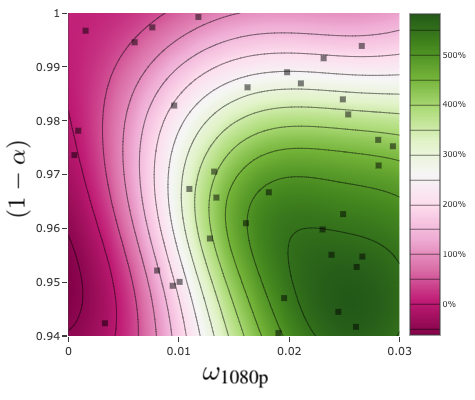}
        \subcaption{\label{fig:video_uploads_contours:quality_preserved} 1080p quality preserved}
    \end{subfigure} %
    \hfill
    \begin{subfigure}{.33\textwidth}
        \includegraphics[width=\columnwidth]{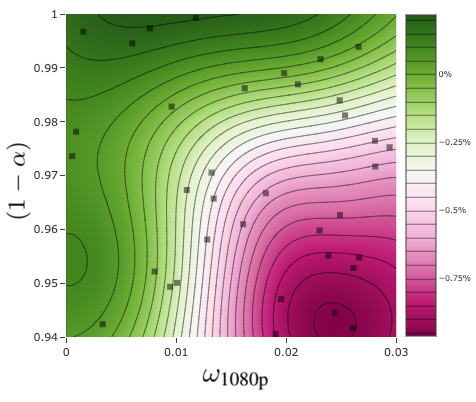}
        \subcaption{\label{fig:video_uploads_contours:reliability} Reliability}
    \end{subfigure}%
    \hfill
    \begin{subfigure}{.33\textwidth}   
    \includegraphics[width=\columnwidth]{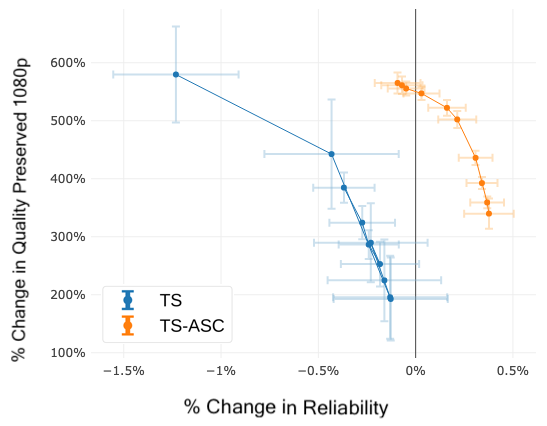}
    \caption{\label{fig:video_uploads_tradeoff} Comparison of quality/reliability trade-off for TS and TS-ASC policies.}
    \end{subfigure}
    \caption{\label{fig:video_uploads_contours} Response surface of mean percent change relative to baseline policy in (a) 1080p quality preserved and (b) reliability, and (c) metric trade-offs for different parameterizations.}
\end{figure}

Figure \ref{fig:video_uploads_contours:quality_preserved} and Figure \ref{fig:video_uploads_contours:reliability} show the modeled response surface for the mean percent change in quality preserved for 1080p and reliability, respectively, relative to the baseline policy. The response surfaces in Figure \ref{fig:video_uploads_contours} demonstrate that $\alpha$ is a powerful lever for constraining in the new policy. When $\alpha$ is close to 0, Figure \ref{fig:video_uploads_contours:reliability} shows no decrease in reliability relative to the baseline policy regardless of the value of reward for a 1080p upload. As $\alpha$ increases, the constraint is relaxed and TS-ASC actions are feasible even with a small relative drop in the expected reliability, which we observe when using a larger reward for  a successful 1080p upload. In correspondence, Figure \ref{fig:video_uploads_contours:quality_preserved} shows the inverse effect in the lower right quadrant. Figure \ref{fig:video_uploads_tradeoff} demonstrates the importance of TS-ASC in finding satisfactory policies: all policies using vanilla TS decrease reliability, whereas there are many parameterizations that lead to TS-ASC policies with increases in both quality and reliability. 

\section{Conclusion}
In this paper, we identify and give formalism to a contextual bandit problem with auxiliary safety constraints, which is ubiquitous in real world applications. We propose a novel algorithm called Thompson Sampling with Auxiliary Safety Constraints (TS-ASC) for learning policies under these conditions, demonstrate its performance empirically, and apply it to a real world video transcoding task, which highlights the necessity of the auxiliary safety constraint for constrained policy optimization. We hope that this paper motivates future research on this type of problem and inspires work on theoretical guarantees in this setting. 

\clearpage

\small
\bibliography{neurips_2019}

\begin{thebibliography}{24}
\providecommand{\natexlab}[1]{#1}
\providecommand{\url}[1]{\texttt{#1}}
\expandafter\ifx\csname urlstyle\endcsname\relax
  \providecommand{\doi}[1]{doi: #1}\else
  \providecommand{\doi}{doi: \begingroup \urlstyle{rm}\Url}\fi

\bibitem[Yauney and Shah(2018)]{yauney18}
Gregory Yauney and Pratik Shah.
\newblock Reinforcement learning with action-derived rewards for chemotherapy
  and clinical trial dosing regimen selection.
\newblock In \emph{Proceedings of the 3rd Machine Learning for Healthcare
  Conference}, volume~85, 2018.

\bibitem[Lazic et~al.(2018)Lazic, Lu, Boutilier, Ryu, Wong, Roy, and
  Imwalle]{lazic18}
Nevena Lazic, Tyler Lu, Craig Boutilier, Moonkyung Ryu, Eehern Wong, Binz Roy,
  and Greg Imwalle.
\newblock Data center cooling using model-predictive control.
\newblock In \emph{Advances in Neural Information Processing Systems 32}, 2018.

\bibitem[Tang et~al.(2013)Tang, Rosales, Singh, and Agarwal]{tang13}
Liang Tang, Romer Rosales, Ajit Singh, and Deepak Agarwal.
\newblock Automatic ad format selection via contextual bandits.
\newblock In \emph{Proceedings of the 22nd ACM international conference on
  Information and Knowledge Management}, 2013.

\bibitem[Bottou et~al.(2013)Bottou, Peters, {n}onero Candela, Charles,
  Chickering, Portugaly, Ray, Simard, and Snelson]{bottou13}
L\'{e}on Bottou, Jonas Peters, Joaquin~Qui\ {n}onero Candela, Denis~X. Charles,
  D.~Max Chickering, Elon Portugaly, Dipankar Ray, Patrice Simard, and
  Ed~Snelson.
\newblock Counterfactual reasoning and learning systems: The example of
  computational advertising.
\newblock \emph{Journal of Machine Learning Research}, 14:\penalty0 3207--3260,
  2013.

\bibitem[Li et~al.(2010)Li, Chu, Langford, and Schapire]{li2010}
Lihong Li, Wei Chu, John Langford, and Robert~E. Schapire.
\newblock A contextual-bandit approach to personalized news article
  recommendation.
\newblock In \emph{Proceedings of the 19th International Conference on World
  Wide Web}. ACM, 2010.

\bibitem[Bastani and Bayati(2015)]{bastani2015online}
Hamsa Bastani and Mohsen Bayati.
\newblock Online decision-making with high-dimensional covariates.
\newblock \emph{Available at SSRN 2661896}, 2015.

\bibitem[Villar and Rosenberger(2018)]{villar2018covariate}
Sof{\'\i}a~S Villar and William~F Rosenberger.
\newblock Covariate-adjusted response-adaptive randomization for multi-arm
  clinical trials using a modified forward looking gittins index rule.
\newblock \emph{Biometrics}, 74\penalty0 (1):\penalty0 49--57, 2018.

\bibitem[Auer(2002)]{auer02}
Peter Auer.
\newblock Using confidence bounds for exploitation-exploration trade-offs.
\newblock \emph{Journal of Machine Learning Research}, 3:\penalty0 285--294,
  2002.

\bibitem[Dani et~al.(2008)Dani, Hayes, and Kakade]{dani2008}
Varsha Dani, Thomas~P Hayes, and Sham Kakade.
\newblock Stochastic linear optimization under bandit feedback.
\newblock \emph{Proc. 21st Annual Conference on Learning Theory (COLT 2008)},
  2008.

\bibitem[Chu et~al.(2011)Chu, Li, Reyzin, and Schapire]{chu11}
Wei Chu, Lihong Li, Lev Reyzin, and Robert~E. Schapire.
\newblock Contextual bandits with linear payoff functions.
\newblock \emph{Journal of Machine Learning Research - Proceedings Track},
  15:\penalty0 208--214, 2011.

\bibitem[Agrawal and Goyal(2013)]{agrawal13}
Shipra Agrawal and Navin Goyal.
\newblock Thompson sampling for contextual bandits with linear payoffs.
\newblock In \emph{International Conference on Machine Learning}, 2013.

\bibitem[Abbasi-Yadkori et~al.(2011)Abbasi-Yadkori, P\'al, and
  Szepesv\'ari]{abbasi13}
Yasin Abbasi-Yadkori, D\'avid P\'al, and Csaba Szepesv\'ari.
\newblock Improved algorithms for linear stochastic bandits.
\newblock In \emph{Advances in Neural Information Processing Systems 24}, page
  2312–2320, 2011.

\bibitem[Filippi et~al.(2010)Filippi, Capp\'e, Garivier, and
  Szepesv\'ari]{filippi13}
Sarah Filippi, Olivier Capp\'e, Aur\'elien Garivier, and Csaba Szepesv\'ari.
\newblock Parametric bandits: The generalized linear case.
\newblock In \emph{Advances in Neural Information Processing Systems 23}, page
  586–594, 2010.

\bibitem[Badanidiyuru et~al.(2014)Badanidiyuru, Langford, and
  Slivkins]{badanidiyuru14}
Ashwinkumar Badanidiyuru, John Langford, and Aleksandrs Slivkins.
\newblock Resourceful contextual bandits.
\newblock In Maria~Florina Balcan, Vitaly Feldman, and Csaba Szepesvári,
  editors, \emph{Proceedings of The 27th Conference on Learning Theory},
  volume~35 of \emph{Proceedings of Machine Learning Research}, 2014.

\bibitem[Agrawal and Devanur(2014)]{agrawal2014bandits}
Shipra Agrawal and Nikhil Devanur.
\newblock Bandits with concave rewards and convex knapsacks.
\newblock In \emph{Proceedings of the fifteenth ACM conference on Economics and
  computation}, pages 989--1006. ACM, 2014.

\bibitem[Amani et~al.(2019)Amani, Alizadeh, and Thrampoulidis]{amani2019linear}
Sanae Amani, Mahnoosh Alizadeh, and Christos Thrampoulidis.
\newblock Linear stochastic bandits under safety constraints.
\newblock \emph{arXiv preprint arXiv:1908.05814}, 2019.

\bibitem[Kazerouni et~al.(2017)Kazerouni, Ghavamzadeh, Abbasi-Yadkori, and
  Roy]{kazerouni17}
Abbas Kazerouni, Mohammad Ghavamzadeh, Yasin Abbasi-Yadkori, and Benjamin~Van
  Roy.
\newblock Conservative contextual linear bandits.
\newblock In \emph{Advances in Neural Information Processing Systems 30}, 2017.

\bibitem[Chapelle and Li(2011)]{chapelle11}
Oliver Chapelle and Lihong Li.
\newblock An empirical evaluation of thompson sampling.
\newblock In \emph{Advances in Neural Information Processing Systems 24}, 2011.

\bibitem[Agrawal et~al.(2017)Agrawal, Avadhanula, Goyal, and
  Zeevi]{agrawal2017thompson}
Shipra Agrawal, Vashist Avadhanula, Vineet Goyal, and Assaf Zeevi.
\newblock Thompson sampling for the mnl-bandit.
\newblock \emph{arXiv preprint arXiv:1706.00977}, 2017.

\bibitem[Riquelme et~al.(2018)Riquelme, Tucker, and Snoek]{riquelme18}
Carlos Riquelme, Georeg Tucker, and Jasper Snoek.
\newblock Deep bayesian bandits showdown: An emprical comparison of bayesian
  deep networks for thompson sampling.
\newblock In \emph{International Conference on Learning Representations}, 2018.

\bibitem[Wilson et~al.(2016)Wilson, Hu, Salakhutdinov, and Xing]{wilson16}
Andrew~Gordon Wilson, Zhiting Hu, Ruslan Salakhutdinov, and Eric~P. Xing.
\newblock Deep kernel learning.
\newblock In \emph{Proceedings of the 19th International Conference on
  Artificial Intelligence and Statistics}, volume~51 of \emph{Proceedings of
  Machine Learning Research}, 09--11 May 2016.

\bibitem[Ng et~al.(1999)Ng, Harada, and Russell]{reward-shaping}
Andrew~Y Ng, Daishi Harada, and Stuart Russell.
\newblock Policy invariance under reward transformations: Theory and
  application to reward shaping.
\newblock In \emph{ICML}, volume~99, pages 278--287, 1999.

\bibitem[Letham et~al.(2019)Letham, Karrer, Ottoni, and Bakshy]{letham2019}
Benjamin Letham, Brian Karrer, Guilherme Ottoni, and Eytan Bakshy.
\newblock Constrained bayesian optimization with noisy experiments.
\newblock \emph{Bayesian Analysis}, 14\penalty0 (2):\penalty0 495--519, 06
  2019.

\bibitem[Rasmussen and Williams(2006)]{rasmussen06}
Carl~Edward Rasmussen and Christopher K.~I. Williams.
\newblock \emph{Gaussian Processes for Machine Learning}.
\newblock The MIT Press, 2006.

\end{thebibliography}
\newpage
\section*{Appendix}
\subsection{Modeling for Video Uploads}
\label{app:modeling}
We use a shared two-layer fully-connected neural network (with 50 and 4 hidden units respectively) as feature extractor to learn a feature representation $\bm \phi_a$ for each context-action feature vector $\mathbf x_a$. We use the learned feature representation as the input to a Gaussian Process (GP) with a linear kernel and a Bernoulli likelihood by placing a Gaussian Process prior over the latent function $\mathbb P(f^{(c)}) = \mathcal N(\mathbf 0, K)$ where $K_{ij} = k(\bm\phi_i, \bm\phi_j) = \bm\phi_i^T \Sigma_p \bm\phi_j$. Since we only have observations of the binary success outcome $y$, we use a Bernoulli distribution for the likelihood of the observed outcomes given the latent function $\mathbb P(y_t|f^{(c)}(\bm \phi_{at}))$, which we can write as: 

$$\mathbb P(\mathbf y|f^{(c)}) = \prod_{t=1}^T \Phi\big(f^{(c)}( \bm \phi_{at})\big)^{y_t}\big(1-\Phi\big(f^{(c)}( \bm \phi_{at})\big)^{1-y_t},$$

where $\Phi$ is the inverse probit link function (Normal CDF). Hence, the joint distribution of the latent function and the observed outcomes is given by

$$\mathbb P(\mathbf y, f^{(c)}) =\prod_{t=1}^T \mathcal{B}(y_t|\Phi(f^{(c)}(\bm{\phi}_{at}))~\mathcal{N}(f^{(c)} | \mathbf 0, K)$$

\citep{rasmussen06}. Since inference over the posterior is intractable with a non-conjugate likelihood, we use variational inference to approximate the posterior and train the model by maximizing the evidence lower bound (ELBO) using stochastic variational inference 1000 inducing points. We use a decoupled training procedure for training the neural network feature extractor and fitting variational GP. The neural network is first trained on end-to-end to minimize the negative log likelihood using mini-batch stochastic gradient descent. Then the last linear layer of the network is discarded and variational GP is fit using learned feature representation.
\end{document}